\documentclass[a4paper]{article}
\usepackage{INTERSPEECH2020}
\usepackage{multirow}
\usepackage{hyperref}
\newcommand{\hyperfootnote}[1][]{\def\ArgI\hyperfootnoteRelay}
\newcommand\hyperfootnoteRelay[2][]{\href{#1#2}{\ArgI}\footnote{\href{#1#2}{#2}}}
\title{Combining speakers of multiple languages to improve quality of neural voices }
\name{Javier Latorre, Charlotte Bailleul, Tuuli Morrill, Alistair Conkie, Yannis Stylianou}
\address {
  Apple
  }
\email{ \{jlatorrechimoto,cbailleul,tuuli\_morril,aconkie,istylianou\}@apple.com}

\begin{document}

\maketitle
\begin{abstract}
%Neural voices based on the sequence-to-sequence paradigm can provide much better quality than traditional methods such as unit-selection of parametric synthesis based on Hidden Markov models. However, this new technology also requires a large amount of speech data to work well.
%Recently, researchers have shown that the lack of  data from the target speaker can be compensated by reusing data from other speakers in the same language.
%Unfortunately, when developing a voice for a new language, there are usually not many additional speakers to select from. 
In this work, we explore multiple architectures and training procedures for developing a multi-speaker and multi-lingual neural TTS system with the goals of a) improving the quality when the available data in the target language is limited and b) enabling cross-lingual synthesis. We report results from a large experiment using 30 speakers in 8 different languages across 15 different locales. The system is trained on the same amount of data per speaker. Compared to a single-speaker model, when the suggested system is fine tuned to a speaker, it produces significantly better quality in most of the cases while it only uses less than $40\%$ of the speaker's data used to build the single-speaker model. In cross-lingual synthesis, on average, the generated quality is within $80\%$ of native single-speaker models, in terms of Mean Opinion Score.  

%we address the questions of whether when training neural TTS systems, the lack of data from one speaker can be compensated by using data from other speakers in different languages, and what architecture and training procedure produces most benefits.  
%The results of a large experiment over 30 speakers in 8 different languages shows that models initially trained with a mixture of speakers and then fine-tuning to each speaker are in general equal or better than  single-speaker models trained with more than the double the amount of speaker-specific data.
%Additionally, we also evaluate to what extent these synthetic voices can speak languages other than the one spoken by the voice talent. Although the results varies largely depending on  the speaker and language combination, on average the MOS of these `foreign' voices is around $80\%$ of the MOS of native single-speaker models.  

\end{abstract}

\section{Introduction}

The quality of synthetic speech has improved dramatically since the development of methods based on neural networks, \cite{wavenet, tacotron2}. 
%At the moment, for some databases and in some applications it is hard to distinguish synthetic speech from real human speech.
However, using this technology requires high computational capacity and large amounts of training data.
%At Apple, we have been able to sort the computational problem so that neural voices can run in our latest iPhones. (CITE PAPER OR REMOVE LINE)
%However, the data requirement remains. 
Several researchers have shown that unlike unit-selection text-to-speech (USEL), neural TTS can compensate for the lack of speech data from the target speaker by adding data from other speakers.
Most of the research published in this respect has used support speakers in the same language as the target. 
However, the most common case when developing TTS voices for a new language is that there are no additional supporting speakers in that new language. 
In that context, the only available options are to record more speakers and/or to use support speakers from different languages.  

In this paper, we  show the results of applying the latter approach on a large-scale experiment involving 30 target speaker in 8 languages across 15 different locales. 
Our goal was to address the following questions: 
a) how effective is it to combine speakers from different languages compared with just training only on the data of the target speaker;  
b) what type of model architecture and training protocol yields the best quality when using multilingual data; and 
c) to which extent can the voices created in this way speak  some of the other languages included in the training data?

In addition to the standard numerical results, we also show the analysis of the most common errors pointed out by the evaluation subjects. 
We believe that the results of these experiments will be useful for researchers and practitioners developing synthetic voices.
  
The structure of the paper is as follows. 
Section \ref{sec:previous_work} reviews the recent literature on using data from other speakers to create new voices and on the application of this method to create polyglot voices. 
Section \ref{sec:architecture} describes the architecture of the models used in the experiment as well as the way in which these models were trained. 
Section \ref{sec:experiments} describes the conditions and results of our experiments. It also shows the analysis of  most commonly mentioned mistakes  for each of the systems. 
Section \ref{sec:discussion} discusses  some of the results and suggests some possible future directions. Finally, in section \ref{sec:conclusions} conclusions are drawn.\footnote{Samples can be found in \url{https://apple.github.io/ml-polyglot_tacotron2_finetuning-samples}}. 
%\hyperfootnote[My website][http://]{www.mywebsite.com}
%\footnote{Samples can be found in \url{https://apple.github.io/polyglot_finetuning}}

\section{Related work}\label{sec:previous_work}

The idea of using  data from other speakers to improve the quality of synthetic speech has been explored extensively  \cite{JunichiSMAPLR2009, CATMultispeakerJournal, Fan2015}.
Although there has been some work in training multi-speaker text-to-wave models  \cite{park2019multispeakerEnd2End}, most of the recent work has been in phone-to-spectrogram. For instance, in  \cite{Latorre2019}  the effect of reducing the amount of data from the target  speaker and compensating for it with  data from other speakers was studied.  The effect of having imbalanced training data was further analized in \cite{luong2019training}. Even more extreme examples were presented in \cite{deng2019modeling}, where only 5 minutes of speech were used to get high quality or even in  \cite{zeroShotSpeakerAdaptationCooper2020} where a single utterance is used.  
When there are not sufficient support speakers, some authors have suggested to artificially expand the number of training speakers \cite{cooper2020speaker} or making use of low quality data \cite{Hu2019}.

Mixing languages has also been widely studied, although in most cases with the goal of creating polyglot voices.  
Within the sequence-to-sequence framework,  \cite{Zhang2019} and \cite{Chen2019} introduced several modifications to allow training polyglot voices using only monolingual speakers. A non sequence-to-sequence model was proposed in  \cite{Himawan2020}.

Even without aiming to create polyglot voices,  using  compensatory data from speakers in other languages is also a potential solution to the lack of data. 
However, this option has received less attention.  
An architecture inspired by the  speaker and language factorisation (SLF) approach \cite{SLF2012} but within the DNN/LSTM framework was  proposed in \cite{Li+2016}. 
Other authors have also shown that mixing data from multiple speakers and languages can yield equal or even better quality than  single speaker models \cite{Fan2016,Yu2016}.
%In \cite{Yu2016,Chen2019b}, multi-language data was used to improve the quality of low-resource languages.
%an automatic phone-mapping between the target-language phones and those of the available data was proposed. 
Finally, in \cite{Indian8Languages},  8 Indian languages were combined directly in a very similar way to the one we suggest here but using a DeepVoice3  \cite{ping2018deep} architecture.  

% Voice conversion based on phonetic posteriograms:  \cite{Nachmani2019}, \cite{zhao2021natural} \cite{Sun+2016}, \cite{sun2020building},

\section{Model training}\label{sec:architecture}
\subsection{Model architecture}
The basic architecture of our models is  Tacotron2 \cite{tacotron2}. The main input is a sequence of phones and punctuation marks and the output is a sequence of 80-dimensional mel-spectrogram features. 
These are computed from speech signals with a sampling rate of 24kHz, using a 25ms analysis window and the Mel filter-bank generated using Librosa Toolkit \cite{librosa}.  
An end-pointing flag was also concatenated with the mel-spectrogram vector to make the final 81-dimensional output vectors. 

The encoder consists of a look-up table that converts the sequence of phone-IDs into a sequence of 512-dimensional vectors,  three 1D-CNNs and one bi-LSTM layers. 
The attention is a stepwise monotonic attention \cite{stepwiseMonotonicAttention}.
On the decoder side, the pre-net consists of two fully connected (FF) layers. 
The decoder itself is formed by two LSTMs followed by one FF layer to decode the mel-spectrograms and another one to generate the end-pointing signal. 
These  mel-spectrograms are finally passed through a post-net module consisting of 5 1D-CNNs. 
The training loss combines the L1 for the end-pointing and the output of the decoder, and an  L2 for the output of the post-net.
 For each output step, two output vectors were generated. 
 \begin{figure}[t]
  \centering
  \includegraphics[width=80mm]{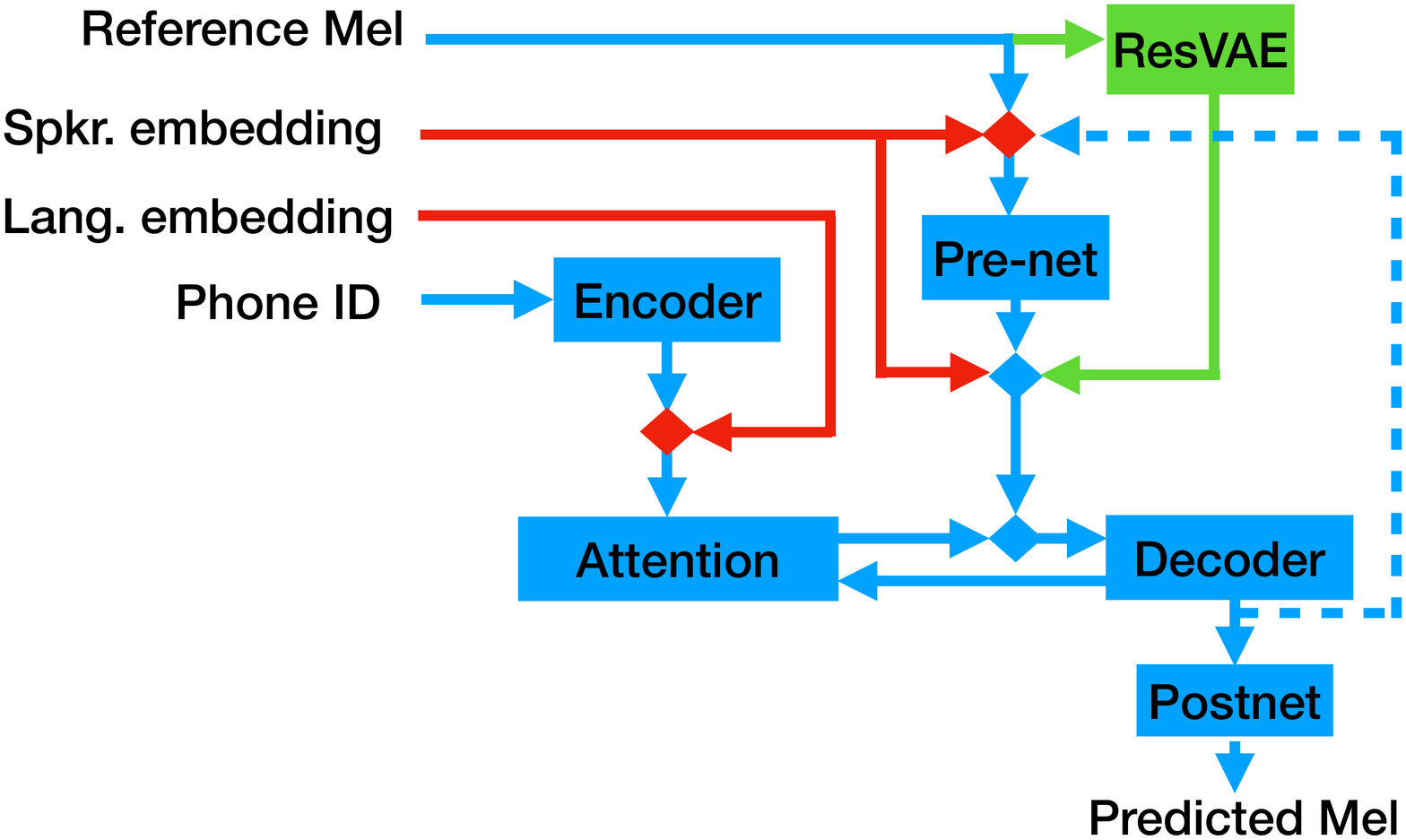}
  \vspace{-10mm}
  \caption{Model architectures. The rhombi indicate concatenation. The dotted line linking the postnet output to the pre-net input is used only at inference time }
  \label{fig:architectures}
  \vspace{-5mm}
\end{figure}

On top of that standard architecture, two variants were built, as depicted in fig. \ref{fig:architectures}
The first variant (in green) consists of  adding a 16-dimensional residual variational auto-encoder (resVAE)  following \cite{Zhang2019}. 
The main goal of the resVAE is to normalise differences between the utterances that cannot be described from the input.
The resVAE consists of 6 2D-CNN layers, each followed by batch normalisation,  a GRU, a common FF layer and 2 additional FF layers, one for the mean and another for the covariance diagonal. 
A sample from this single Gaussian distribution is then concatenated at the input of the decoder. 
As usual, an additional loss factor for the KLD w.r.t a diagonal Gaussian  was added.
During inference the resVAE network is bypassed and a constant 0-vector is used instead. 

The second variant (in red) is the addition of speaker and language embeddings. 
The speaker embedding (SE) consists  of a 128 dimensional d-vector obtained from a speaker verification model  \cite{Hu2019}.  
One advantage of using speaker embeddings versus one-hot is that we can have different values for each utterance. 
Unfortunately, the SE of each utterance also contains information about the acoustics of that utterance \cite{Soumi2020}. 
To avoid this and simulate something akin to a VAE, a single Gaussian model  of  the embeddings of each  speaker was computed and sampled during training. 
At inference, the mean of the Gaussian was used.  
The language embeddings (LE) are 32 dimensional vectors obtained from a one hot encoding of the locale associated with each speaker. 
%For some languages there are multiple locales, so for example, there are different embeddings for British and American English.
 
We  experimented with different ways of adding SE and LE. 
For SE, we found that the best option is to insert it both at the input of the decoder, concatenated with the output of the attention and with the output of the pre-net as in \cite{Xue2019}. 
This configuration yields the best results in terms of quality, voice similarity to the target speaker and voice consistency when synthesising mix-lingual sentences. 
For LE, the best option was to  concatenate it  with the output of the encoder before the attention. 
Concatenating LE at the beginning of the encoder or after the attention made  the models' training unstable. 
In any case, the effect of LE was almost negligible, presumably because the phonetic sequence itself already contains enough information about the language.

Previous internal evaluations on models with SE but without fine tuning showed a preference for  adding the resVAE. 
For that reason, all our models with  SE also include resVAE.  
In some initial models we also included a domain adversarial NN (DANN) loss against the identification of the speaker from the encoder outputs as suggested in \cite{Zhang2019}. 
Although DANN provided some good results when mixing only 2-3 languages with at least 4 speakers each \cite{Soumi2020}, it introduced instability when we added languages for which only two speakers were available.

\subsection{Training procedure}
Models that do not include any speaker information need to be fine-tuned in order to get a stable voice. 
Models that include SE can be used either directly, as in \cite{Latorre2019}, or they can also be fine-tuned. 
%Previous work \cite{Hu2019} showed that fine-tuned models tend to perform better than models with only SE. 
%Preliminary experiments with single-language multi-speaker models showed no improvement when SE-models were further fine-tuned to the already included the target speaker. 
%However, we wanted to test whether this was also true when mixing data in different languages. 
All the base models were trained on exactly the same data. 
For the fine-tuning to each target speaker we used exactly the same utterances of that speaker that were used as part of the base-model training.   
We didn't consider experiments in which an existing model was fine-tuned to an unseen speaker because if the data for the new speaker is available, it can always be mixed with the existing speakers to create a new  base model.  

All models were trained on a single GPU with a batch size of 16. 
We used the Adam optimiser \cite{kingma2017adam} with 0.9 and 0.999 for beta1 and beta2, respectively, an initial learning rate of 0.001, 4000 warm-up steps and ``Noam decay scheme"  \cite{vaswani2017attention}.
The seed models were trained for  2.5 million steps and then fine-tuned for another 0.5 million steps.
The non fine-tuned seed model with speaker embeddings  was further trained  up to 4.5 million steps.  
The  systems that were finally evaluated  are shown in Table \ref{tab:architectures_summary} .
\begin{table}[tb]
\begin{center}
\caption{Evaluated models}
\vspace{-2mm}
\begin{tabular}{ c |  c | c | c  | c }\label{tab:architectures_summary}
	 System &  Fine Tuned (FT) & resVAE & SE+LE & \#total steps\\ 
	 \hline
	  FT & Yes &No & No   & 3x$10^6$\\
	  FTres & Yes & Yes & No & 3x$10^6$ \\ 
	  FTresSE & Yes & Yes & Yes  &3x$10^6$\\ 
	  resSE & No & Yes & Yes & 4.5x$10^6$\\ 
  \hline
\end{tabular}
\end{center}
\vspace{-9mm}
\end{table}

\subsection{Normalisation of the phonetic transcriptions}
We normalised the transcriptions across all locales to share a single unified language-agnostic set of phones based on XSAMPA \cite{XSAMPA}.
Previous experiments had shown that in crosslingual synthesis complex phones such as diphthongs, nasalized vowels, syllabic consonants and affricates, tend to get confused and the synthesis only produces half of the phone.
To avoid this problem, we split such complex phones. In this way, diphthongs were split into two vowels, affricates into a closure with no audible release plus a fricative, syllabic consonants into the consonant preceded by schwa, and nasalized vowels into a vowel followed by a velar nasal consonant.  

Syllabic stress marks were also added to the vowels of the stressed syllables for all languages. 
It should be noted that for inlingual synthesis, (in which the spoken language is the same language as that of the target voices) most languages do not need explicit stress marks, especially those languages for which stress is not phonemic.
However, we found that in crosslingual synthesis, (which is when the synthesised utterances were in a language other than that of the target voices) the lack of stress marks caused serious intelligibility problems, even in languages which are supposed to have no phonemic stress, such as French. In crosslingual synthesis, the voices tended to apply the stress pattern of its own language, e.g., Spanish voices speaking French tended to put the stress in the penultimate syllable. 
Such changes of the stress patterns made the parsing of the prosodic words very difficult and thus, affected the intelligibility of the utterances.   

\section{Experiments}\label{sec:experiments}
We ran two subjective evaluations, one for inlingual synthesis and another for crosslingual synthesis. 
%In addition to the systems described in Table \ref{tab:architectures_summary}, we also evaluated single speaker models (`SingSpkr') trained from scratch on all the available data of each speaker. 
%In the crosslingual case, `SingSpkr' corresponds to models trained on native speakers in the target language and is used as the upper anchor.
All the evaluations were 5 points mean opinion score (MOS) tests conducted via crowdsource on each respective locale. 
The question asked was ``How do you rate the overall quality of the voice?". 
Each utterance was evaluated by 15 different subjects and no subject was allowed to judge more than 360 samples. %It appears from the worker files that it was 356 but there are a couple cases of slightly higher so I said 360%Tuuli, do you remember this number?, 
With these settings, the total number of listeners per voice was around 120 for the inlingual experiments and 140 for the crosslingual one.
 
%Tuuli, could you describe here the MOS normalization protocol and how we compute the statistical significancy?
%from Tuuli: I've added this, not sure how clear you want to be in which cases the stats were by system and which by voice...but this method applied in both cases
For each evaluation, raw scores were normalized by z-scoring by subject. Mixed effects linear regression models were fitted to the data with subjects and items (sample content/sentence) as random effects and the synthesis method/voice as the fixed effect. T-tests for pairwise contrasts for each pair of voices/systems provided estimated p-values (with Bonferroni correction for the number of contrasts).
\subsection{Data}
%\vspace{-2mm}
The models were trained on 30 proprietary voices consisting of  
two speakers for 15 different locales in 8 languages: Australia, India, Ireland, South Africa, UK and US for English; Mexico and Spain for Spanish; Canada and France for French, Brazil for Portuguese, and Denmark,  Germany,  Italy and The Netherlands for their respective main languages. 
From each speaker we used 8500 utterances  randomly selected from the total corpus, which on average corresponds to 7.73 hours/speaker. This amount of data corresponds on average to 37\% of the data used to train the single speaker (SingSpkr) models. 
\subsection{Vocoder}
%\vspace{-2mm}
In all the experiments, we used speaker-dependent waveRNN neural vocoders \cite{KalchbrennerWaveRNN}. 
The same vocoder trained on all the data was used for each voice across all the models. 
The reasons for this are: a) we only wanted to evaluate differences in the acoustic model and, b) there exist proposals for universal waveRNN that work for both seen and unseen speakers \cite{UniversalWaveRNN2019,Paul2020}

\subsection{Inlingual synthesis}\label{sec:Inlingual}
%\begin{table*}[tb]
%\begin{center}
%\caption{Inlingual synthesis.}
%\begin{tabular}{ c | c | c | c | c | c | c | c  }\label{tab:inlingual_results}
%	  & Recordings & USEL & SSM  & FT & FTres & FTresSE & resSE \\ 
 % \hline
%    MOS (mean;median)  & 4.40;4.43& 3.91;3.48& 4.04;4.10& 4.12;4.15& 4.14;4.14& 4.14;4.14& 4.03;3.99\\ 
%    normMOS (mean;median)  & 0.62;0.59& 0.37;0.37& 0.51;0.52& 0.53;0.53& 0.54;0.53& 0.54;0.54& 0.50;0.50\\
%  \hline
%    \#voices significantly $>$ single spkr & -  & 1& -&11&13&13&4\\
%    \#voices significantly  $<$ single spkr  & - &26 &-&1&2&1& 11\\
%    \#voices  significantly $\approx$ single spkr &  - & 1 &-&16&13&14&13\\
 %   \hline
 % \end{tabular}
%\end{center}
%\end{table*}
\begin{figure}[tb] 
  \centering
  \includegraphics[width=85mm]{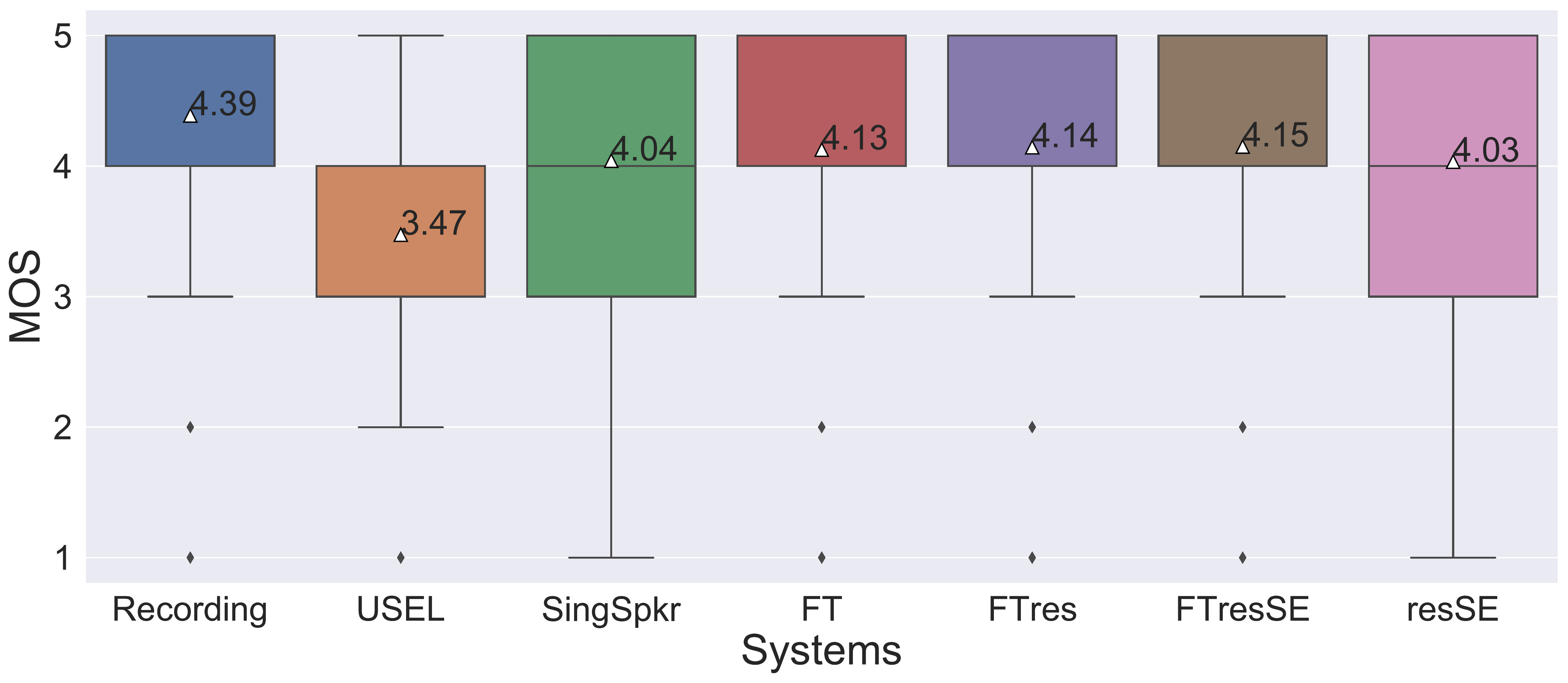}
  \vspace{-5mm}
  \caption{MOS scores across all voices for inlingual synthesis}
  \label{fig:inlingual_boxplot}
  \vspace{-1mm}
\end{figure}

For each of the 15 locales, 150 utterances were evaluated with each of the 2 speakers' voices. 
In addition to the systems described in Table \ref{tab:architectures_summary}, we also evaluated SingSpkr models with the same architecture of FT models, trained from scratch on all the available data of each speaker. 
The sentences were the same for all the systems but not necessarily the same for both speakers.  
In order to provide anchors,  each evaluation included 50 recorded utterances from each of the target voice talents as the high anchor and  the same 150 evaluation sentences\footnote{For 1 of the 15 locales we used 75 instead of 150 USEL utterances } generated by a hybrid unit selection system (USEL) \cite{AppleUSEL} as the lower one.

Figure  \ref{fig:inlingual_boxplot} shows the box plot with the summary of the results  across all voices.
On average, all the fine-tuned models outperformed the SingSpkr models.  
The average difference between the models is around 0.1 MOS scores. Note that this is by using less than 40\% of the target speaker data of the SingSpkr models.  %I think we talked about changing this section and removing this reference to the score(?) 
%Javier that was more for the cross-lingual part, I think. 
Obviously, there are  variations depending on the voice.  A  voice-by-voice analysis is provided in Table \ref{tab:inlingual_significance_voices}. 
This result confirms that for most voices any of the fine-tuned models perform equal or better than the SingSpkr models. 
By contrast, resSe was found to be significantly worse than SingSpkr for 11 voices and only better for 4, even though  both systems appear to be identical in fig. \ref{fig:inlingual_boxplot}. Our results confirm  those reported  in \cite{Indian8Languages} for premium voices with 15+ hours of data.
Finally, we did not find  any significant differences among the 3 fine-tune approaches, although both FTres and FTresSE seem to be  marginally better than FT, presumably due to their higher capacity.    
 \begin{table}[tb]
\begin{center}
\caption{Number of voices significantly different from SingSpkr models in inlingual synthesis.}
\vspace{-2mm}
\begin{tabular}{ c |  c | c | c | c | c  }\label{tab:inlingual_significance_voices}
	  & USEL & FT & FTres & FTresSE & resSE \\ \hline
    better & 1&11&14&14&4\\
    equal &  1 &18&14&15&15\\
    worse  & 28 &1&2&1& 11\\
    \hline
  \end{tabular}
\end{center}
\vspace{-8mm}
\end{table}

\subsection{Crosslingual synthesis and evaluation}
Our main purpose was to create a base model from which new voices for new languages can be created rapidly. However, given that the seed models are trained on multiple languages, we were curious to know to which extent the fine-tuned models still retained some multilingual capacity.
Evaluating each of the 30 voices over the 7 non-native languages would have been ideal, but also very costly.
For that reason, we  evaluated only the  non-native voices when synthesising 4 different foreign languages, American English (en-US), Mexican Spanish (es-MX),  France French (fr-FR) and Germany German (de-DE).
%is de-DE meant to be added because it says 4 languages?
%It depends on whether we get the results on time 
For each target language 50 utterances from each  of the non-native voices  were evaluated.  Voices  in the same main language but from a different locale  were not considered. %, i.e., British voices were not evaluated for American English. 
To avoid conflating the differences between native/non-native speakers with those between synthetic/natural speech, we only included as upper anchor 50 utterances generated by the native SingSpkr voice in the target language.  These SingSpkr voices are the same as the ones described in Sec. \ref{sec:Inlingual}. 
%To reduce the number of different voices/systems in a single  evaluation, the stimuli were split according to the gender of the voice. 
To reduce the number of different voices/systems in a single evaluation, the stimuli were split into two groups: one for the voices with the lower median fundamental frequency (F0) and another for the voices with higher median F0 from each locale.
This yields a total of 8 independent MOS evaluations.
In total,  the number of individual voices evaluated on  each experiment were 10 for English, 14 for Spanish and French and 15 for German.  Subjects were not warned that they were going to listen to foreign accented speech.

%Note that since the values of male and female for each row should not be compared directly as they were done in different MOS tests. 
%To get a general idea, across all voices and languages, we computed the ratio between the average MOS of the  native systems and the crosslingual systems on an  utterance basis. The distribution of these ratios can be seen in the boxplot of figure \ref{fig:crosslingual_boxplot_ratios}. In the figure the upper tail of the ratios has been clipped at 1.5 but it extends in some cases above  3.0
Table  \ref{tab:crosslingual_significancy} shows for how many of the 8 evaluations the MOS difference between models were significant, and Table \ref{tab:crosslingual_results_by_language} reports the average MOS for each combination of target-language and voice-locale.
In general, all systems' performance is very similar. For most voices the non fine-tuned system resVAE is usually better than the fine-tuned ones. This result is not surprising since  fine-tuned models tend to ``forget" previous knowledge. However,  with the exception of the low-pitch voices in French those differences were not significant. 
Among the fine-tuned models,  FTres and FTresSE were better than FT on average, probably because of  the higher capacity introduced by the additional resVAE. However, the addition of the speaker embedding does not seem to provide any advantage when the model is fine-tuned.  

Despite these differences, the combination of voice and target language has a much stronger impact over the speech quality than the model type. As shown in Table \ref{tab:crosslingual_results_by_language}, some combinations achieve scores around 4.0 while  others fall below 3.0.
\begin{table}[tb]
\begin{center}
\caption{Number of model comparisons across the 8 crosslingual evaluations in which the MOS difference was significant}\label{tab:crosslingual_significancy}
\vspace{-2mm}
\begin{tabular}{c| c | c |c}
Systems & \#$1^{st}$ better &  \#$2^{nd}$ better & \#No diff.\\\hline
resSE vs. FT & 1& 0 &7\\
resSE vs. FTres &1 & 0&7\\
resSE vs. FTresSE &1 & 0&7\\
FT vs. FTres&0 &3&5\\
FT vs. FTresSE& 2& 6&0\\
FTres vs. FTresSE&2 &2 &4\\
\hline
\end{tabular}
\end{center}
\vspace{-8mm}
\end{table}

\begin{table}[!ht]
\begin{center}
\caption{Crosslingual MOS  per locale. The numbers in the target language column are the average MOS of the two `Native SingSpkr' voices in that language.}\label{tab:crosslingual_results_by_language}
\vspace{-2mm}
\begin{tabular}{ p{1.1cm} | p{1.1cm}  | p{0.5cm} | p{0.6cm} | p{0.95cm} | p{0.6cm}  }
    	  \shortstack[c]{Target\\ language}&\shortstack[c]{Speaker\\ locale}&FT&FTres&FTresSE&resSE\\\hline

%	  Target      &  Speaker  & Native         &\multirow{2}{*}{FT} &\multirow{2}{*}{FTres}&\multirow{2}{*}{FTresSE}&\multirow{2}{*}{resSE}  \\
%	   language&  locale &  SingSpkr  &       			     &              			 &                      		&			               \\ \hline
    	  \multirow{9}{*}{\shortstack[c]{American\\English\\4.2}} &  da-DK &  3.77&    \textbf{3.84}  &  3.78               & 3.7\\
				               				 &  de-DE  &3.88 &    3.88  &   3.96                       & \textbf{4.01}\\
    		          		       				& es-ES &3.66   & 3.74  &   3.77                      & \textbf{3.78}\\
	  				       				& es-MX &3.81&   3.82   &   3.81                        & \textbf{3.93} \\ 
					       				&  fr-CA   &3.75    & 3.82   &   3.81                                    & \textbf{3.93}\\
					       				&  fr-FR  &3.62  &  3.69  &   3.72                           & \textbf{3.79}\\
	  				       				& it-IT       &3.72 & 3.69    &   \textbf{3.76}                    & \textbf{3.76} \\
					       				&  nl-NL    &3.84  &  3.87   &  3.85                      & \textbf{3.9}\\
					       				&  pt-BR    & 3.68  &  3.71   &  3.81                         & \textbf{3.88}\\
					       \hline
	\multirow{13}{*}{\shortstack[c]{France\\French\\4.35}}  &  da-DK     &   3.08   &  3.25          &   3.26                  & \textbf{3.34}\\
					     				    &  de-DE     &  3.77    &   3.84       &    3.84                 & \textbf{3.86}\\
				               				& en-AU      & 3.08     &  3.11             &  3.2                   & \textbf{3.42} \\
				              				 & en-GB   & 3.28     &  3.33             &  3.3                   & \textbf{3.51} \\
				               				& en-IE        & 3.29     &  3.32             &  \textbf{3.49}    & 3.46 \\
				               				& en-IN       &  3.46    &  3.43             &  3.5                   & \textbf{3.7} \\
				               				& en-US      &  3.27    &  3.29             &  3.29                   & \textbf{3.57} \\
				               				& en-ZA       & 3.25     & 3.47              &  3.47                   & \textbf{3.69} \\
    		          		       				& es-ES      & 3.37     & 3.46              & 3.48                    & \textbf{3.67}\\
				               				& es-MX    & 3.55     &  3.53             &  3.58                   & \textbf{3.7} \\ 
					       				& it-IT      &  3.62    &   3.54            & 3.66                    & \textbf{3.81} \\
					       				&  nl-NL      &  3.18    &  3.52          &  3.32                   & \textbf{3.66}\\
					       				&  pt-BR   &   3.27   &  3.4          &  3.56                   & \textbf{3.72}\\
					       \hline
     	  \multirow{13}{*}{\shortstack[c]{Mexican\\Spanish\\4.45}}&  da-DK&   2.88& \textbf{3.08}  & 2.87 &2.67\\
					      						     &  de-DE& 3.3   & 3.35  & 3.18 &\textbf{3.38}\\
				               						    & en-AU &  2.94	& 2.91 & 2.9 &\textbf{3.04}\\
				               						    & en-GB & 3.15  & 2.93  & 3.02&\textbf{3.13}\\
				               						   & en-IE & 3.03   	& 2.98  & 3.07 &\textbf{3.21}\\
				               						 & en-IN & 3.34   	& \textbf{3.45}  & 3.15&3.32\\
				               						& en-US &  \textbf{3.24}  & 3.14  & 3.12 &3.14\\
				               						& en-ZA&  2.95  & 3.08 & 3.07 &\textbf{3.2}\\
					       						&  fr-CA &   3.23   & 3.05  &3.09  &\textbf{3.46}\\
					      						 &  fr-FR  &   3.41   & 3.45  & 3.37&\textbf{3.47}\\
	  				       						& it-IT    &  3.95 	&  \textbf{4} & 3.92 &3.95\\
					      						 &  nl-NL &   3.03   & 3.08  & 2.85&\textbf{3.08}\\
					       						&  pt-BR  &  3.63  &  3.62 & 3.61 &\textbf{3.64}\\
						\hline
 				       
    	  \multirow{14}{*}{\shortstack[c]{Germany\\German\\3.96}} 	&  da-DK    & 3.27   &  \textbf{3.32}  &  3.3   & 3.3\\
				               							& en-AU   &  3.47    & 3.58   &  3.57   & \textbf{3.63}\\
												& en-GB   & 3.51    &  3.51  & 3.55    & \textbf{3.66}\\
				           						      & en-IE  &  3.57     &  3.7  &   3.66  & \textbf{3.82}\\
				              							 & en-IN   &  3.6       &  3.67  & 3.6    & \textbf{3.77}\\
				               							& en-US  & 3.62     &  \textbf{3.66}  & 3.52    & \textbf{3.66}\\
				               							& en-ZA  &  3.66    &  3.75  &  3.69   & \textbf{3.8}\\
    		          		       							& es-ES   &  2.88   & 3.12   & 3.09    & \textbf{3.44}\\
				              							 & es-MX  &   3.2      & 3.29   & 3.46    & \textbf{3.55}\\
					       							&  fr-CA    &  3.16    & 3.25   & 3.28    & \textbf{3.67}\\
					       							&  fr-FR    &  3.34     &  3.4  &  3.42   & \textbf{3.59}\\ 
	  				      							 & it-IT       &  3.16     & 3.18   & 3.12    & \textbf{3.57}\\
					      							 &  nl-NL    &  3.43    &  3.55  &  3.51   & \textbf{3.65}\\
					       							&  pt-BR     &  2.8       &  2.97  & 3.27    & \textbf{3.55}\\\hline
        \multicolumn{2}{c|}{Total}&									  3.39	& 3.44& 3.44 & \textbf{3.57}\\\hline

					 \hline

    \hline
  \end{tabular}
\end{center}
\vspace{-10mm}
\end{table}

\subsection{Analysis of the comments}
In  inlingual synthesis, the main problems noted  were in terms of pauses (either misplaced or too few), pace (usually too fast), unnatural intonation, and audio quality deterioration. These problems seem to affect  more the resSE model. Word stress also seems to be sometimes slightly misplaced in some languages. In the resSE model, some non-phonemic distinctions are also less accurately predicted, e.g., the Italian trill is sometimes chosen instead of the flap.

In crosslingual synthesis, the foreign accent of a voice is usually well identified, but in some cases deemed too pronounced to the extent of impeding intelligibility, especially when in combination with insufficient pausing and fast pace.
%Degraded intelligibility is not caused only by non-native accents, however. Other factors such as incorrect or insufficient pausing and inconsistent or fast pace also contribute to it.
We also notice a degraded audio quality, affecting some voices more than others, with some occasional ``blabber''. Intonation contours are sometimes incorrect and sometimes deemed as monotone. 
In terms of pronunciation, the model without fine tuning seems to retain less accent and produces a more accurate approximation of the target language phones. 
This effect is notable, for example, with the American English  rhotic and the French voices: the model without fine tuning being the closer to the English alveolar approximant (although getting inaudible in word final position or pre-consonantical position), and other models having a pronunciation closer or identical to the French rhotic. For most voices, lexical stress seems to be placed correctly.

In some language pairs, some phonemic distinctions are lost. For example, the Spanish trill/flap pair is not always maintained when synthesising with French, English, or German voices. 
The English phone /h/ is often dropped in the synthesis with the French voices. Actually, human French speakers often do drop that phoneme. However, it contributes to the impression of strong foreign accent as  more proficient speakers would tend to realise it.
Another factor contributing to the impression of strong foreign accent is that intonation and some phonological phenomena are ported to the target language. For instance, word final rhotic is  dropped by British English voices, and sometimes French voices insert liaison in Spanish.
It is also interesting to note that some American English subjects  expected a genuine non-native accent.  For example, they expected /t/ or /d/  instead of flaps in the  French, Portuguese and German voices.
\begin{figure}[tb] 
\centering
  \includegraphics[width=85mm]{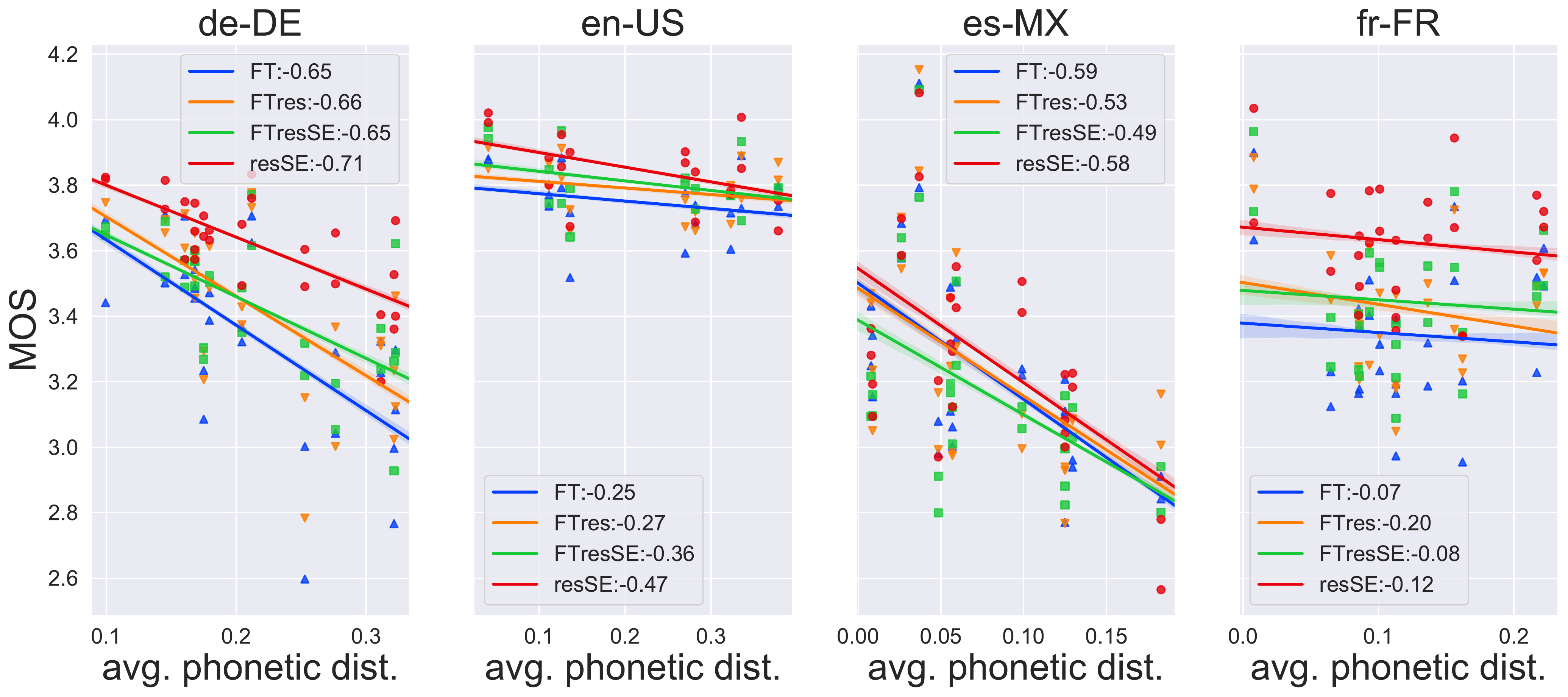}
  \caption{Average crosslingual MOS per voice w.r.t average phonetic distance between voice and language }
  \label{fig:MOS_vs_phoneticCoverage}
\vspace{-5mm}
\end{figure}
\section{Discussion}\label{sec:discussion}
\subsection{Differences by  language}
There are two interesting observations from the crosslingual evaluation.  
The first one is the large variation in the MOS  depending on the combination of target voice/language as shown in Table \ref{tab:crosslingual_results_by_language}. 
One possible explanation for this is that the phonetic differences between the voice's language and the target language matters.  
Figure \ref{fig:MOS_vs_phoneticCoverage} shows the average MOS\footnote{The MOS values have been shifted so that the average MOS across all the samples of the two evaluation groups of each target language are the same}  of the voices/systems in the crosslingual evaluation with respect to the average phonetic distance between the speaker data and the test sentence of the target computed as 
\begin{equation}
AvgPhoneDist=\sum_{\forall t \in T} P( t | T)\min_{\forall s \in S} dist(e_t,e_s)
\end{equation}
where $T$ and $S$ are the sets of unique phones in the test utterances of the target language and in the target speaker data, respectively; $e_t$ and $e_s$ denote the phone embeddings for $t$ and $s$ taken from the look-up-table  of the resSE model, and $dist$ is the cosine similarity function. Note that for all $t\in S$ the minimum distance is 0. 
%the average minimum cosine distance between the embeddings of the phones in the target language test utterances and the embeddings of the phones in the training data of the speaker. Since the phone set is shared across languages, if the target phone was found in the speaker's data, the distance is 0. The phone embeddings were taken from the look-up table of the resSE model.
% computed as the average cosine distances between the phone embeddings of the phones in the test utterances and the phones in the 8500 training utterances of each speaker, with the phone embeddings taken from then \textit{resSE} model. 

Figure \ref{fig:MOS_vs_phoneticCoverage} shows that the impact of the phonetic distance depends on the target language. 
For instance, Mexican subjects penalised foreign accented voices heavily,  even when the average phonetic distance is small. 
On the contrary, for French subjects other factors seem to be more important. 
%It appears that a number of factors seem to influence the results. In the case of the Danish voice, audio quality issues seem to affect intelligibility. 
For example, for the British, Australian and Irish English voices, the factors that  produce the most negative impact are very unnatural and strongly pronounced intonation,  unnatural parsing of groups of words and pace which generally affected intelligibility. For  the British English voices, the intelligibility is also affected by the porting of the non-rhotic character of British English to French: final   /r/ are often dropped and the quality and length of the previous vowel is modified. The pronunciation of French diaeresis also seems problematic in terms of intelligibility, being realised as a diphthong (as in "pays" for instance).
On the other hand, German, Italian, Portuguese, and Spanish voices were preferred in terms of general intelligibility,  even though the intonation was found too monotone, the pauses sometimes incorrectly placed or missing, and the foreign accent too strong.

For American English, MOS is also strongly correlated with the average phonetic distance, but mainly due to the smaller dispersion. Otherwise, the curves are flatter than for German or Spanish. 
This links with the second observation which is that the  average MOS for American English is higher than for the other languages.  
One explanation of that higher score is that an average of 7\% of the 8500 training utterances of the non-English voices were in English, with another 13\% having at least one English word. 
The  English proficiency of the voice talents varied greatly, from fully bilingual to very accented.  
Moreover, the English utterances in the training data of many voices were transcribed using the phones of the voice's language, which might be the reason for the relatively larger phonetic distances for en-US.
Still, that English data seems to have contributed to improve the synthesis of English utterances with non-English voices. 
 Another possible explanation for the higher MOS for American English  may be that subjects in that locale (and to some extent in France French too) are more used to listening to foreign accents than their  Mexican or German counterparts and therefore, have a larger tolerance for them. Further experiments are needed to confirm which hypothesis is correct. 

%Another possible explanation is that voices with a more varied  phoneme inventory are better at producing "foreign" phones. 
%One finding supporting this hypothesis is that the ratios between the native and the foreign systems among France and Germany subjects were higher than among  Mexican subjects. Another is that less likely pairs, like Spanish-Danish, were usuall the ones with the worse scores.

%To investigate this, we analyse the relationship between the phonetic coverage of each target voice over the evaluation utterance and their average MOS  
%Figure \ref{fig:MOS_vs_phoneticCoverate} shows this for male and female tests. The numbers on the legend indicate the correlation score between these average MOS and phonetic coverage.(CAN WE SAY SOMETHING ABOUT THIS GRAPH? IF NOT REMOVE)
\vspace{-1mm}
\subsection{Pauses}
One of the most commented problems for inlingual synthesis was  errors with pausing. 
Since the model does not include any explicit pause predictor, or part-of-speech tagging, the pause prediction depends entirely on the phonetic transcription and punctuation marks. 
In single-language models, the network  might be able to perform some level of syntactic parsing, for example identify the most common function words. 
In a multilingual framework, this is harder because the same phonetic sequence might also correspond to a content word in a different language. 
But also, different languages have different rules regarding the punctuation. 
So, whereas in some languages it is used mostly to indicate pausing, in others they have a more  grammatical function. 
These kinds of differences are hard to disambiguate by just looking at the phone sequence.    
Including a LE was expected to help with such language-dependent issues. However, simply concatenating a global LE at the input of the attention didn't work.

\section{Conclusions}\label{sec:conclusions}
This paper confirms that data from speakers in other languages can be used to compensate for the lack of target speaker data.
We have presented a large-scale experiment on building neural TTS models by mixing speech from 30 speakers of 15 different locales in 8 different languages.  
The results show that for the vast majority of voices, fine-tuning a multi-lingual and multi-speaker model produces equal or better quality than single-speaker models trained with more than 2.5 times the amount of speaker-specific data.  

An evaluation of these models synthesizing speech in a language different from that of the target speaker has confirmed that the models also preserve good multilingual capability. 
On average, the MOS on these models in a crosslingual scenario is around 80\% of the MOS obtained by inlingual single-speaker native voices.  Although this may not be enough for a general  stand-alone voice in that language, it is sufficient for code-switching.  
Our results showed that although non fine-tuned voices are marginally better for crosslingual synthesis,  for inlingual synthesis they are generally significantly worse than the fine-tuned ones. 
Finally, we have presented a qualitative  analysis of the main problems identified by subjects during the inlingual and crosslingual evaluations.   

%\section{Acknowledgements}

\newpage
\bibliographystyle{IEEEtran}

\bibliography{mybib}

% \begin{thebibliography}{9}
% \bibitem[1]{Davis80-COP}
%   S.\ B.\ Davis and P.\ Mermelstein,
%   ``Comparison of parametric representation for monosyllabic word recognition in continuously spoken sentences,''
%   \textit{IEEE Transactions on Acoustics, Speech and Signal Processing}, vol.~28, no.~4, pp.~357--366, 1980.
% \bibitem[2]{Rabiner89-ATO}
%   L.\ R.\ Rabiner,
%   ``A tutorial on hidden Markov models and selected applications in speech recognition,''
%   \textit{Proceedings of the IEEE}, vol.~77, no.~2, pp.~257-286, 1989.
% \bibitem[3]{Hastie09-TEO}
%   T.\ Hastie, R.\ Tibshirani, and J.\ Friedman,
%   \textit{The Elements of Statistical Learning -- Data Mining, Inference, and Prediction}.
%   New York: Springer, 2009.
% \bibitem[4]{YourName17-XXX}
%   F.\ Lastname1, F.\ Lastname2, and F.\ Lastname3,
%   ``Title of your INTERSPEECH 2020 publication,''
%   in \textit{Interspeech 2020 -- 20\textsuperscript{th} Annual Conference of the International Speech Communication Association, September 15-19, Graz, Austria, Proceedings, Proceedings}, 2020, pp.~100--104.
% \end{thebibliography}

\end{document}